\documentclass[10pt,twocolumn,letterpaper]{article}

\usepackage{multirow}
\usepackage[pagenumbers]{cvpr} 

\usepackage[dvipsnames]{xcolor}

\newcommand{\todo}[1]{{\color{red}#1}}

\usepackage{epsfig}
\usepackage{graphicx}
\usepackage{amsmath}
\usepackage{amssymb}
\usepackage{multirow}
\usepackage{graphics}
\usepackage{threeparttable}
\usepackage{color}
\usepackage[normalem]{ulem}
\usepackage{float}
\usepackage{amsfonts}
\usepackage{bm}
\usepackage{bbm}
\usepackage{enumitem}
\usepackage{algorithmic,algorithm}
\usepackage{natbib}
\usepackage{subcaption}
\usepackage{algorithm}
\usepackage{array}
\usepackage{colortbl}
\usepackage{pifont}
\usepackage{booktabs}
\usepackage{mathtools}
\usepackage{siunitx}
\usepackage{caption} 
\usepackage{xcolor}
\usepackage{subcaption} 
\usepackage[nopar]{lipsum}
\usepackage{listings}
\usepackage[export]{adjustbox}
\usepackage{makecell}

\usepackage{mathtools}
\usepackage{siunitx}
\usepackage[nopar]{lipsum}
\usepackage{listings}

\usepackage{xargs}                      
\usepackage{xcolor}
\usepackage{wrapfig} 
\newcommandx{\yiyuan}[2][1=]{\todo[linecolor=red,backgroundcolor=red!25,bordercolor=red,#1]{#2}}
\newcommandx{\xiangyu}[2][1=]{\todo[linecolor=blue,backgroundcolor=blue!25,bordercolor=blue,#1]{#2}}
\newcommandx{\kaixiong}[2][1=]{\todo[linecolor=OliveGreen,backgroundcolor=OliveGreen!25,bordercolor=OliveGreen,#1]{#2}}
\newcommandx{\improvement}[2][1=]{\todo[linecolor=Plum,backgroundcolor=Plum!25,bordercolor=Plum,#1]{#2}}
\definecolor{myy}{RGB}{126,95,0}
\definecolor{mygray}{gray}{.9}
\definecolor{Gray}{gray}{0.9}
\definecolor{bblue}{RGB}{30,80,120}
\definecolor{mygray1}{gray}{.7}
\definecolor{ggray}{RGB}{127,127,127}
\definecolor{defaultcolor}{gray}{.9}
\definecolor{dark-gray}{gray}{0.20}

\definecolor{mygreen}{HTML}{39b54a}
\newcommand{\pub}[1]{{\color{dark-gray}{\tiny{[{#1}]}}}}

\newcolumntype{x}[1]{>{\centering\arraybackslash}p{#1pt}}
\newcolumntype{y}[1]{>{\raggedright\arraybackslash}p{#1pt}}
\newcolumntype{z}[1]{>{\raggedleft\arraybackslash}p{#1pt}}

\newlength\savewidth

\definecolor{cvprblue}{rgb}{0.21,0.49,0.74}
\usepackage[pagebackref,breaklinks,colorlinks,citecolor=cvprblue]{hyperref}
\usepackage{cuted}
\usepackage{capt-of}
\def\paperID{****} 
\def\confName{****}
\def\confYear{****}

\title{\textit{InteractiveVideo}: User-Centric Controllable Video Generation with Synergistic Multimodal Instructions}

\author{
~~~~ {Yiyuan Zhang}$^{1*} $ 
~~~  {Yuhao Kang}$^{2*}$
~~~~ {Zhixin Zhang}$^{2}$\thanks{Equal Contribution} \\
~~~ {Xiaohan Ding}$^{3}$
~~~ {Sanyuan Zhao}$^{2}$
~~~ {Xiangyu Yue}$^{1}$ \\
\textsuperscript{1} Multimedia Lab, The Chinese University of Hong Kong \\
\textsuperscript{2} Beijing Institute of Technology
\quad
\textsuperscript{3}Tencent AI Lab~~~ \\
{\tt\small 
	 yiyuanzhang.ai@gmail.com,
    kangyuhao@bit.edu.cn,
    xyyue@ie.cuhk.edu.hk}\\
 {\small \url{https://invictus717.github.io/InteractiveVideo}}
	}

\begin{document}
\maketitle

\begin{strip}\centering
	\vspace{-19mm}
	\includegraphics[width=\textwidth]{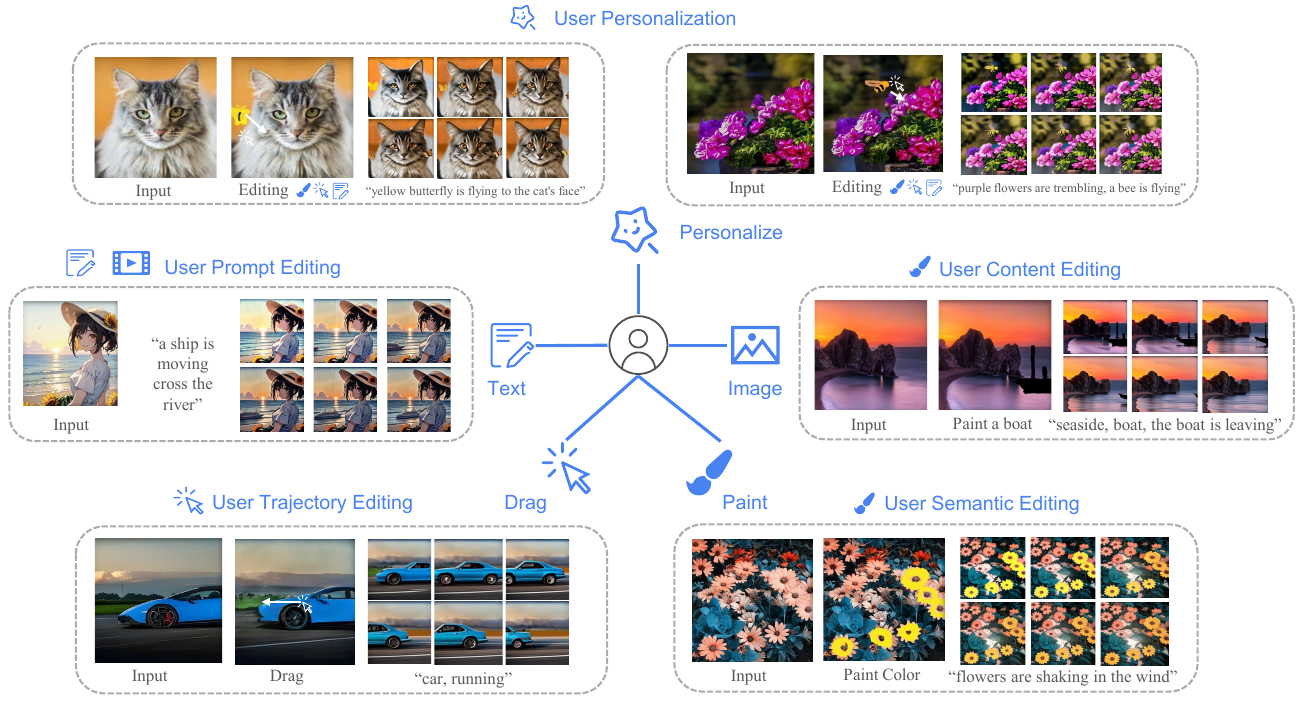}
	\captionof{figure}{\textbf{Interactive Video Generation} We propose a user-centric framework that effectively synergizes users' multimodal instructions. Users can easily edit key components in the video generation process, leading to high-quality video and increased user satisfaction. 
		\label{fig:teaser}}
  \vspace{-1.5mm}
\end{strip}

\begin{abstract}
 We introduce ``InteractiveVideo'', a user-centric framework for video generation.	Different from traditional generative approaches that operate based on user-provided images or text, our framework is designed for dynamic interaction, allowing users to instruct the generative model through various intuitive mechanisms during the whole generation process, 
e.g. text and image prompts, painting, drag-and-drop, etc.
 We propose a Synergistic Multimodal Instruction mechanism, designed to seamlessly integrate users' multimodal instructions into generative models, thus facilitating a cooperative and responsive interaction between user inputs and the generative process.
 This approach enables iterative and fine-grained refinement of the generation result through precise and effective user instructions.  
 With \textit{InteractiveVideo}, users are given the flexibility to meticulously tailor key aspects of a video. They can paint the reference image, edit semantics, and adjust video motions until their requirements are fully met. Code, models, and demo are available at \url{https://github.com/invictus717/InteractiveVideo}.
\end{abstract}

\clearpage
\begin{strip}
    \centering
    \includegraphics[width=0.98\linewidth]{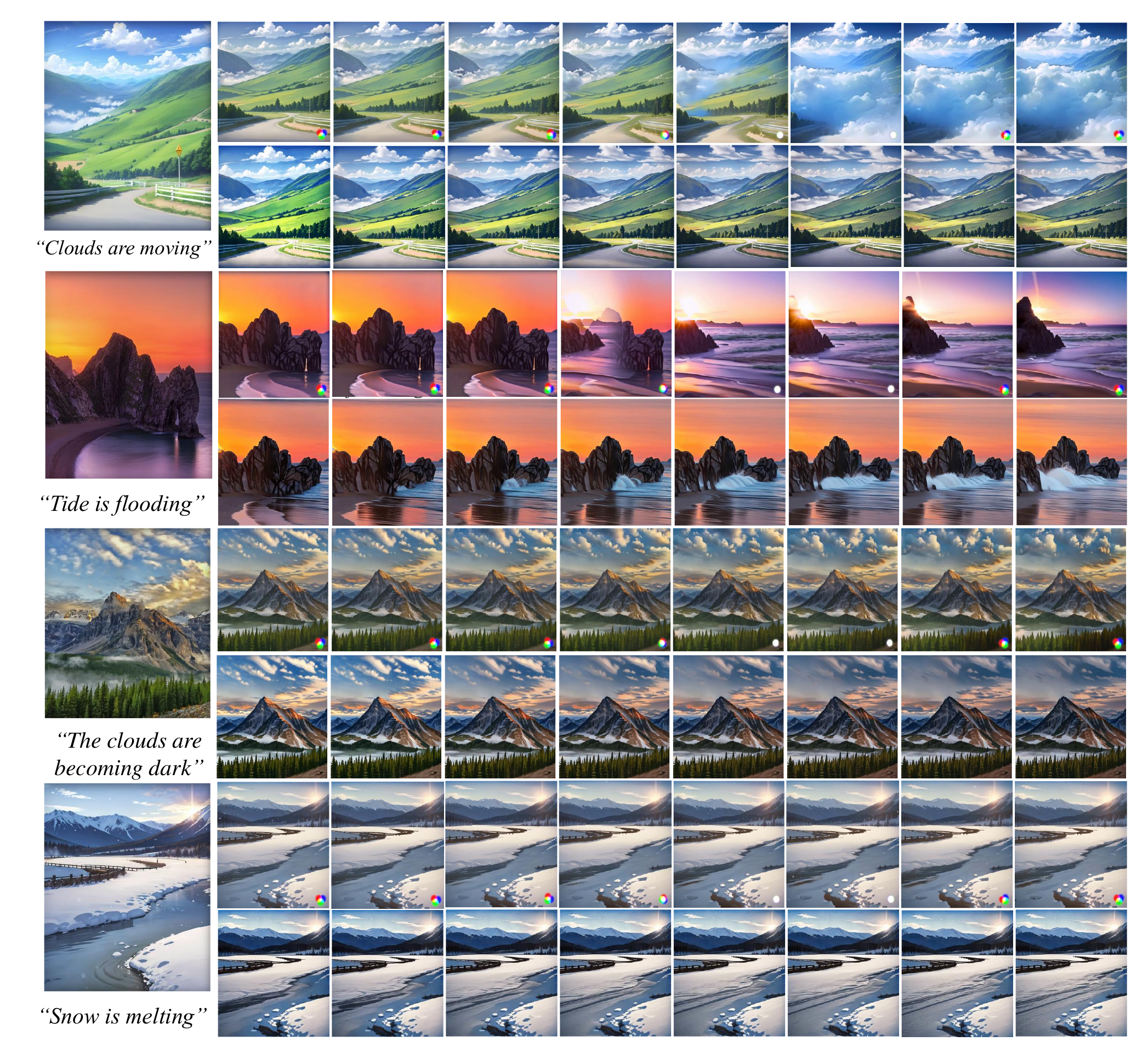}
    \captionof{figure}{Comparison between Gen-2 and \textit{InteractiveVideo}. For each case, the first row is the generation results of Gen-2, and the second row is our results. (More comparison results with Pika Labs, I2VGen-XL~\cite{zhang2023i2vgen-xl}, and Gen-2 can be found in Appendix Figures~\ref{fig:6}, \ref{fig:7}, and \ref{fig:8}.)}
    \label{fig:cmp}
\end{strip}

\section{Introduction} ~\label{sec:intro}
Video generation has attracted significant attention due to its promising future in the AI-Generated Content field~\cite{yin2023dragnuwa, ruan2023mm-diffusion, chen2023videocrafter1, khachatryan2023text2video-zero, xing2023dynamicrafter} and its potential to enhance the efficiency of movie creation and serve as a new infrastructure technology for the film industry~\cite{skorokhodov2022stylegan-v,ho2022imagenvideo,zhou2022magicvideo, esser2023gen-1}. The advancement of diffusion models~\cite{sohl2015Diffusion_model,ho2020denoising_DDPM,zhang2022gddim,song2020DDIM} has infused the field of video generation with new potential~\cite{chen2023controlavideo,chen2023videocrafter1,ruan2023mm-diffusion}. The success of Gen-1~\cite{esser2023structure}, MagicVideo~\cite{zhou2022magicvideo}, and Align your latents~\cite{blattmann2023align} has significantly inspired further exploration of high-quality visual content generation for videos.

As the capabilities of video generation models improve, user expectations for the generated videos are concurrently elevated, leading to an increased demand for videos that accurately meet their specific requirements. Existing video generation models typically utilize a reference image, known as the \textit{image condition}, and a textual description, referred to as the \textit{text condition}, as inputs. 
Enhancing video generation foundation models by making them larger, more advanced, and more sophisticated could potentially fulfill user requirements more effectively by enhancing the understanding of image and text conditions, thereby producing videos of superior quality.
However, our objective is to empower existing video generation models to more accurately fulfill user requirements from a different angle - by equipping models with the capability to interpret complex, multidimensional human instructions. This approach is driven by the observation that the current conditioning mechanisms (image and text) fall short of capturing the full spectrum of user intentions.
 \textbf{1}) The text condition may not be informative enough. Even though existing video generation models support long and detailed text prompts, it is difficult to precisely depict complex video motions and dynamics using only text. 
 As a result, it becomes challenging for generative models to fully interpret the intended video content. 
 \textbf{2}) The conditional image does not contain temporal information. The absence of optical flow and temporal consistency can easily lead to the introduction of unsatisfactory artifacts in the video generation process. 
 Moreover, \textbf{3}) there is a significant demand from users for the customization of videos, which entails the intuitive manipulation of video contents, semantics, and motions. In response to these challenges, we propose a novel approach that improves the ability of existing video generation models to better understand human intentions and generate videos guided by more detailed and multifaceted human instructions.

Recently, remarkable progress in large language models has drawn widespread attention across the community~\cite{chang2023survey}. One key to the success of large language models is learning from human feedback through reinforcement learning~\cite{christiano2017deep_RLHF, leike2018scalable_agent_align,ouyang2022InstructGPT, stiennon2020summerize_HF} which significantly improves the performance of language models and leads to superior generation results. Pioneers in the visual content generation field have also introduced human feedback to generate high-quality images~\cite{xu2023ImageReward}. 
Nonetheless, the intricacy, diversity, and level of control required for video generation far surpass those needed for single-image generation, making it a highly significant yet relatively underexplored challenge.

To address these challenges, we propose \textit{InteractiveVideo}, a user-centric video generation framework that empowers users to actively participate in the generation process through 
multimodal instructions, enabling control over video content, semantics, and motions. Users can customize a video through various manipulations such as painting and dragging, as illustrated in Figure~\ref{fig:teaser}. More specifically, we propose a Synergistic Multimodal Instruction mechanism that empowers generative models to interpret and act upon users' editing and revision instructions across various facets, such as video content, regional semantics, object motion, subjects, and the overall dynamics of the video. In our framework, we capture user interactions in the form of image, text, motion, and trajectory prompts, and we incorporate these user instructions as independent conditions of probabilistic models. As a result, \textit{InteractiveVideo} is a training-free framework that can be easily and flexibly applied to different fundamental generative models. It is worth noting that our proposed framework seamlessly integrates with existing generative models and practical techniques, such as Stable Diffusion~\cite{rombach2022stable_diffusion}, DreamBooth~\cite{ruiz2023dreambooth}, and LoRA~\cite{hu2021lora}, thus expanding the video generation capabilities with our interactive framework.

In this framework, user interactions are involved through four distinct types of instructions which can be employed independently or collaboratively to effectively guide the video generation process. The four types of instructions are: 
\begin{itemize}
    \item Image Instruction: the image condition or prompt for image-to-video generation.
    \item Content Instruction: a textual description of the visual elements and the painting edits of the user to control the video content.
    \item Motion Instruction: a textual description specifying the desired movements and dynamics of elements within the video.
    \item Trajectory Instruction: user-defined motion trajectories for specific video elements, expressed through interactions such as dragging.
\end{itemize} 

By incorporating these detailed and multidimensional human instructions, we can generate videos that better align with the unique preferences and requirements of users.

We compare our \textit{InteractiveVideo} with the advanced video generation solutions, \eg  Gen-2\footnote{https://research.runwayml.com/gen2}, I2VGen-XL~\cite{zhang2023i2vgen-xl}, and Pika Labs. Comparison results in Figure~\ref{fig:cmp}, \ref{fig:6}, \ref{fig:7}, \ref{fig:8} show the superiority of \textit{InteractiveVideo} with higher quality, better flexibility, and richer controllability. Our \textit{InteractiveVideo} paves the way for a novel paradigm in visual content generation, integrating user interactions to enable highly customized video generation. This empowers users to effortlessly obtain high-quality videos they desire through intuitive manipulation and effective interactions.

In summary, our contributions are as follows:
\begin{itemize}
    \item \textbf{Framework Deisgn}: we propose a novel interactive framework that empowers users to precisely control video generation by intuitive manipulations. 
    \item \textbf{Generation Algorithm}: we propose a Synergistic Multimodal Instructions mechanism, which integrates user prompts as probabilistic conditions and enables interaction without the need for additional training.
    \item \textbf{High-quality Video Generation}: our generation results demonstrate superiority over state-of-the-art video generation methods, including Gen-2, I2VGen-XL~\cite{zhang2023i2vgen-xl}, and Pika Labs.
\end{itemize}

\newpage

\section{Related Work}~\label{sec:rel}
\begin{figure*}[t]
    \centering
    \includegraphics[width=0.91\linewidth]{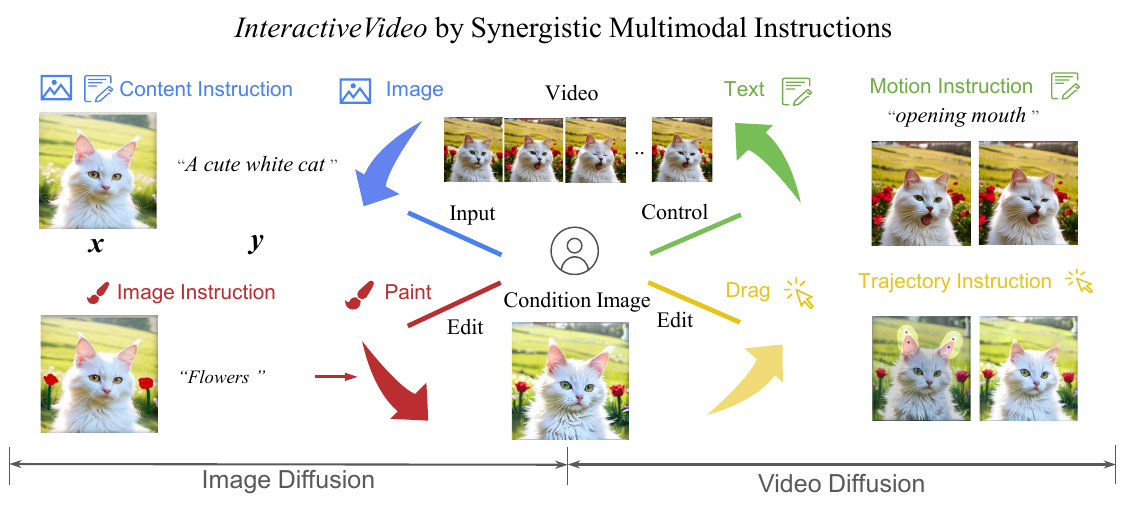}
    \caption{\textbf{Framework Illustration}. In \textit{InteractiveVideo}, users can utilize multimodal instructions to interact with generative models on video content, motion, and trajectory.}
    \label{fig:2}
\end{figure*}
\subsection{Video Generation}

Initial attempts in video generation primarily leveraged Generative Adversarial Networks (GANs) \cite{pan2017TGANs-C, li2018video_GAN_VAE, villegas2022phenaki_GAN, skorokhodov2022stylegan-v, hong2022cogvideo, yu2022video_implicit_GAN, wu2022nuwa, fu2023tell_me} and Variational Autoencoders (VAEs) \cite{li2018video_GAN_VAE, mittal2017sync-draw, yan2021videogpt}. These methods, however, faced considerable challenges in effectively modeling the intricate spatio-temporal dynamics necessary for text-driven video generation, leaving the problem largely unsolved. Subsequent innovations shifted towards diffusion models \cite{sohl2015Diffusion_model,ho2020denoising_DDPM,zhang2022gddim,song2020DDIM} to enhance diversity and fidelity in video outputs \cite{ho2022imagenvideo, ho2022video_diffusion_models, zhang2023show-1, blattmann2023align, yu2023PVDM, zhou2022magicvideo, esser2023gen-1, yin2023dragnuwa, ruan2023mm-diffusion, chen2023videocrafter1, khachatryan2023text2video-zero, xing2023dynamicrafter} and to scale up pre-training data and model architecture \cite{singer2022make-a-video, hong2022cogvideo, wu2021godiva, ho2022imagenvideo,zhang2023meta,ding2023unireplknet}. Recent efforts have introduced spatio-temporal conditions \cite{chen2023controlavideo, 2023videocomposer, yin2023dragnuwa, esser2023gen-1}, for instance, through VideoComposer \cite{2023videocomposer}, Gen-1 \cite{esser2023gen-1} and DragNUWA \cite{yin2023dragnuwa}. These methods aim to provide a more controlled generation process but still encounter constraints in achieving flexible and user-satisfied video synthesis.

\subsection{Models Guided by Human Feedback}

The idea of learning from human feedback, initially investigated in reinforcement learning and agent alignment contexts \cite{christiano2017deep_RLHF, leike2018scalable_agent_align}, was subsequently applied to large language models \cite{ouyang2022InstructGPT, stiennon2020summerize_HF}. This approach has significantly improved the generation of textual outputs that are helpful, honest, and harmless. In the field of visual content generation, particularly in video generation and editing, a similar goal is pursued. \cite{xu2023ImageReward, wu2023HPSv1, lee2023AlignT2I} demonstrates the great potential of human guidance for the visual content generation field.

These works collectively highlight the growing trend of incorporating human feedback in various forms of generative models, extending its utility from text-based to visual content generation. However, learning from human feedback for video generation remains under-explored owing to its complicated elements of motion, subjects, and spatial-temporal dynamics. We aim to fill this gap, providing a training-free and user-friendly solution for elevating existing video generative models with effective human guidance and generating more user-satisfying and higher-quality videos.

\section{Methodology}

\subsection{Preliminary}~\label{sec:method:3.1}

As shown in Figure~\ref{fig:2}, \textit{InteractiveVideo} realizes controllable video generation with two generative pipelines based on latent diffusion models - \textbf{1}) the text-to-image (T2I) pipeline $\mathcal{P}_{img}$ and \textbf{2}) the image-to-video (I2V) pipeline $\mathcal{P}_{video}$. The framework outputs a video containing $N_F$ frames $\{\boldsymbol{v}_1, \boldsymbol{v}_2, \cdots, \boldsymbol{v}_{N_F}\}$. We denote the Image Instruction by $\boldsymbol{x} \in \mathbb{R}^{C \times H\times W}$, the Content Instruction by $y$, the Motion Instruction by $y^\prime$, and the Trajectory Instruction by $r$. More specifically, the Trajectory Instruction is represented by start and end points and region masks, which indicate the desired moving trajectories of specific objects. The whole pipeline can be formulated as 
\begin{equation}
    \label{eq:ivideo_pipe}
    \{\boldsymbol{v}_1, \boldsymbol{v}_2, \cdots, \boldsymbol{v}_{N_F}\} = \mathcal{P}_{video}(\mathcal{P}_{img}(\boldsymbol{x},y),y^\prime, r).
\end{equation}

In practice, we may implement $\mathcal{P}_{img}$ with any off-the-shelf T2I model as long as it takes a text condition and an image condition as inputs. We use $\tilde{\boldsymbol{x}}$ to denote its generated image, \ie, the intermediate image, which is the input to the I2V model.

We then use $\tilde{\boldsymbol{x}}$ as the image condition of the I2V pipeline and the Motion Instruction $y^\prime$ as the text condition. We may use any off-the-shelf I2V diffusion models which require image and text conditions. Let $\mathcal{E}$ be the image encoder of the I2V model, $\boldsymbol{z}_0=\mathcal{E}(\tilde{\boldsymbol{x}})$ be the corresponding latent code, $\epsilon_t$ be the predicted noise at step $t$, the classic  (\ie, interaction-free) video denoising process can be denoted with
\begin{equation}
    \label{eq:i2v}
    \boldsymbol{z}_t = \sqrt{\bar{\alpha}_t} \boldsymbol{z}_0 + \sqrt{1-\bar{\alpha}_t} \cdot \epsilon_t, 
\end{equation}
where $\bar{\alpha}_t$ is a parameter related to the variance schedule~\cite{ho2020denoising}.

\subsection{Synergistic Multimodal Instructions}~\label{sec:method:3.2}

We control the video diffusion process with users' multimodal instructions via \textit{altering the predicted noise according to the users' operations}. Conceptually, with $R$ denoting the function that changes $\epsilon_t$ according to users' operations, our interaction-controlled video diffusion process can be represented by
\begin{equation}
    \label{eq:i2v-trajectory}
    \boldsymbol{z}_t = \sqrt{\bar{\alpha}_t} \boldsymbol{z}_0 + \sqrt{1-\bar{\alpha}_t} \cdot R(\epsilon_t).
\end{equation}
The proposed concrete implementation for the function $R(\cdot)$ is realized by treating user interactions as denoising residuals. Since the intermediate image $\tilde{\boldsymbol{x}}$ is utilized as the condition image of $\mathcal{P}_{video}$, it is seen as the ``interface'' between user and video generation model. Consequently, our framework empowers users to interact with the target video by introducing their interactions as new generation conditions of the video denoising process. 

Specifically, we transform users' operations into denoising residuals to eventually control the video diffusion process. Formally, in the video denoising process, assume the original intermediate image is $\tilde{\boldsymbol{x}}$ and the corresponding latent code is $\boldsymbol{z}_0=\mathcal{E}(\tilde{\boldsymbol{x}})$. Once the user has operated on the image (\eg, painted some lines or set some trajectories\footnote{Painting and trajectory drawing affect the intermediate image in different ways. The former makes a difference on the intermediate image through the T2I pipeline as it changes the very beginning Image Instruction. The latter moves the handle points within the specified region to the target points and changes the optical flow of the intermediate image.}), the intermediate image changes accordingly, and we denote the resultant intermediate image as $\tilde{\boldsymbol{x}}^\prime$ and the corresponding latent code becomes $\boldsymbol{z}_0^\prime=\mathcal{E}(\tilde{\boldsymbol{x}}^\prime)$. We use $\boldsymbol{z}_0^\prime$ to predict the noise in the video diffusion process. Formally, let $t$ be the time step, $\epsilon_t$ be the noise predicted with $\boldsymbol{z}_{t-1}$ and $\epsilon_t^\prime$ be the noise predicted with $\boldsymbol{z}_{t-1}^\prime$, the noise we use is given by 
\begin{equation}
    \label{eq:interact_impact}
    \hat{\epsilon}_t = \lambda \cdot \epsilon_t + (1-\lambda)\cdot\epsilon_t^\prime \,,
\end{equation}
where $\lambda$ is a hyper-parameter to balance the learned noise residual and human instructions.
Then we use $\hat{\epsilon}_t$, instead of the original $\epsilon_t$, in the denoising process to generate the eventual video.

\begin{figure*}[t]
    \centering
    \includegraphics[width=0.89\linewidth]{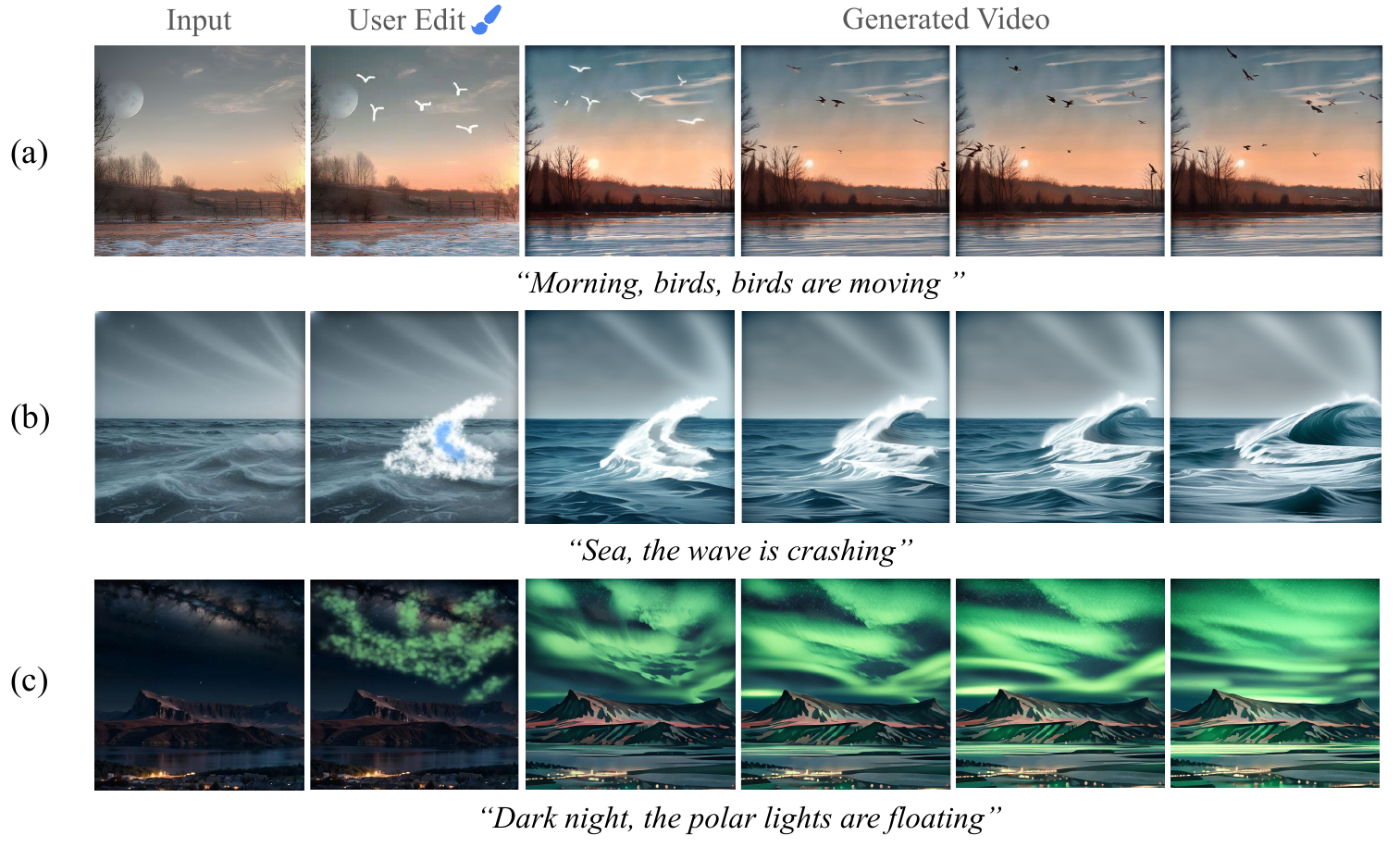}
    \caption{\textbf{Video Content Manipulation} with \textit{InteractiveVideo}. In (a), (b), and (c), we present the content manipulation by adding birds, waves, and polar lights. Then, these added objects are driven in the whole video. We use these results to show the flexibility of our framework for video content creation.}
    \label{fig:3}
\end{figure*}
\begin{figure*}[t]
    \centering
    \includegraphics[width=0.90\linewidth]{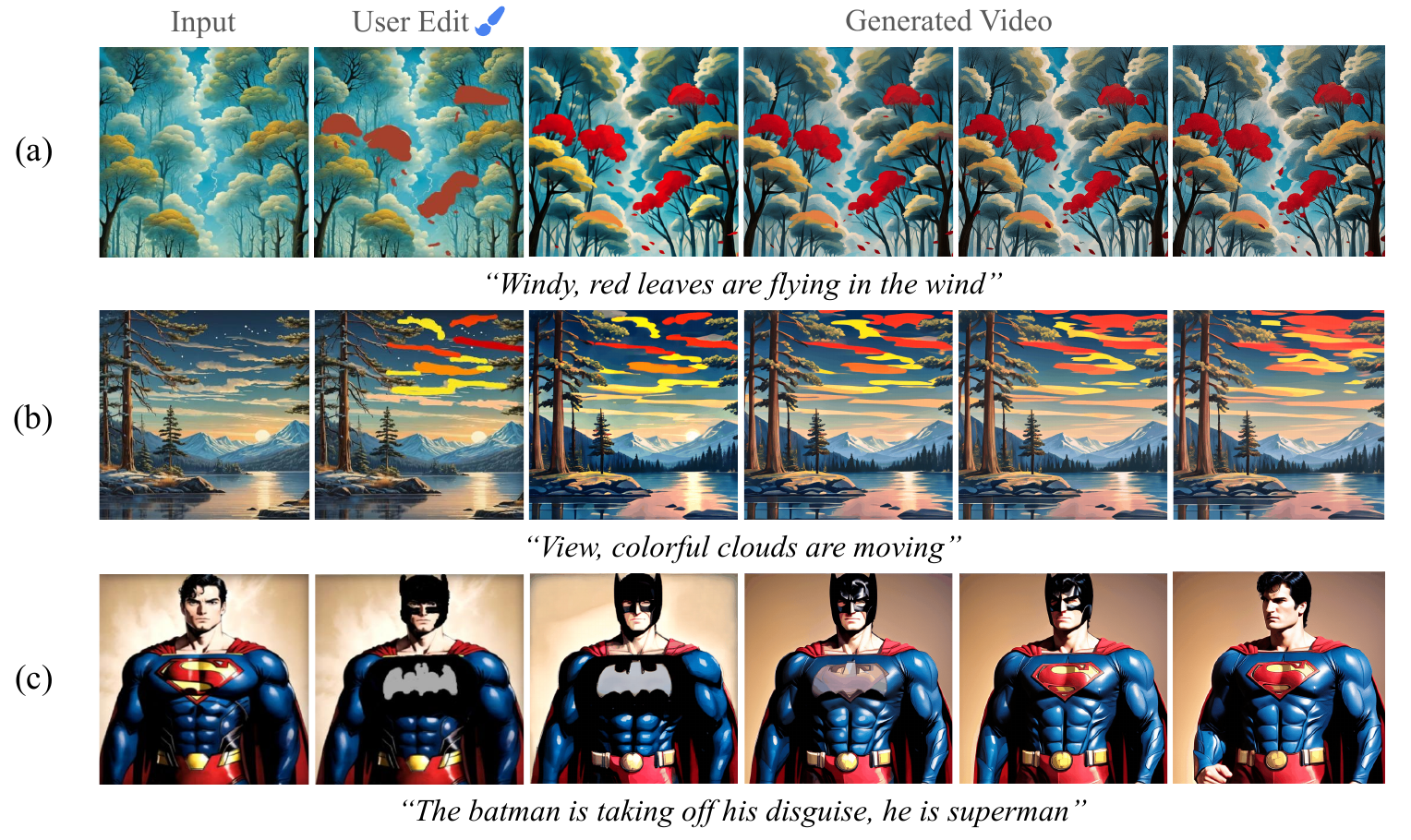}
    \caption{\textbf{Fine-grained Video Editing} with \textit{InteractiveVideo}. In (a), (b), and (c), we perform fine-grained \textbf{regional} semantic editing on changing colors and appearances of specific objects, These results show the outstanding controllability of our framework for video generation.}
    \label{fig:4}
\end{figure*}

Note that after the user operations on the image, the human-crafted discrepancy may affect the temporal coherence of the resulting video. This is because the user operations may have deviated the intermediate image from the distribution on which the I2V model was trained (\eg, the user has drawn a twisted yellow curve to create a sun in the sky, which is unusual in the training data of the I2V model). To solve this problem, upon the completion of the video diffusion process, we post-process the resultant video following AnimateDiff~\cite{guo2023animatediff}. Every single frame is aligned with the intermediate image via a Group Normalization~\cite{wu2018group} layer, a SiLU~\cite{hendrycks2016gaussian} activation, and a 2D convolutional layer adopted from AnimateDiff or PIA~\cite{zhang2023pia}, as such structures are found to generalize well to our common distribution produced by users' typical operations. Specifically, the eventual $i$-th video frame $\boldsymbol{v}_i^\prime$ can be computed as: 
\begin{equation}
    \label{eq:synergy}
    \begin{aligned}
        \boldsymbol{v}_i^\prime = \mathtt{Conv2D}(\mathtt{SiLU}(\mathtt{GroupNorm}(\boldsymbol{v}_i - \tilde{\boldsymbol{x}}))) \,.
    \end{aligned}  
\end{equation}

\section{Experiments}~\label{sec:exp}

In this section, we present features of the \textit{InteractiveVideo} framework including personalization (\S~\ref{sec:exp:4.1}), fine-grained video editing (\S~\ref{sec:exp:4.2}), and precise motion control (\S~\ref{sec:exp:4.3}). Besides, we also conduct quantitative analysis (\S~\ref{sec:exp:quan}) on generation quality and user study on the satisfaction rates with our framework. Then, we demonstrate the generation efficiency (\S~\ref{sec:exp:efficiency}) of our framework.

\subsection{Personalizing a Video}~\label{sec:exp:4.1}
Existing methods~\cite{guo2023animatediff,2023videocomposer,esser2023gen-1} have made significant progress in the animation of static images into videos. However, these methods are limited to animating objects or scenes already present in the original static images, and encounter difficulties when it comes to generating a video with objects or scenes absent from the referenced images. In other words, existing methods have limited ability to control video content, especially when users want to add or animate previously unseen objects or scenes.

With \textit{InteractiveVideo}, we enable video content manipulation by incorporating abundant elements. In Figure~\ref{fig:3}, we demonstrate that our framework supports the users to customize the video content freely. For example, we use a brush to paint sketches of birds, waves, and polar lights in Figure~\ref{fig:3} (a), (b), and (c), respectively. The added objects are seamlessly integrated and animated throughout the entire video. Seen from the following frames, \textit{InteractiveVideo} enables users to create a video of satisfactory temporal consistency even though the referenced image does not directly contain the objects. Meanwhile, such cases also demonstrate the versatility and adaptability of our framework in creating diverse and engaging video content, highlighting its potential for a wide range of applications in content creation and editing.

\subsection{Fine-grained Video Editing}~\label{sec:exp:4.2}

Another significant limitation of current generation methods is the challenge of performing precise regional editing. During the generation process, models have difficulty interpreting natural language references such as ``left'', ``right'', ``up'', and ``down''. This makes it hard to accurately edit regional semantics, which is crucial for user experience.

Fortunately, \textit{InteractiveVideo} overcomes this limitation by enabling intuitive manipulation in the intermediate image. As illustrated in Figure~\ref{fig:4} (a), it is difficult for users to edit the color of a specific tree or control the color of a particular cluster of falling leaves using existing methods. In contrast, our framework allows users to perform fine-grained semantic editing on any region. For example, after the editing process, the trees in Figure~\ref{fig:4} (a), clouds in Figure~\ref{fig:4} (b), and the logo in Figure~\ref{fig:4} (c) can be easily modified. The generated videos are of high quality, featuring realistic motion, appropriate light reflection, and visually appealing textures.

\begin{figure}[ht]
    \centering
    \includegraphics[width=0.98\linewidth]{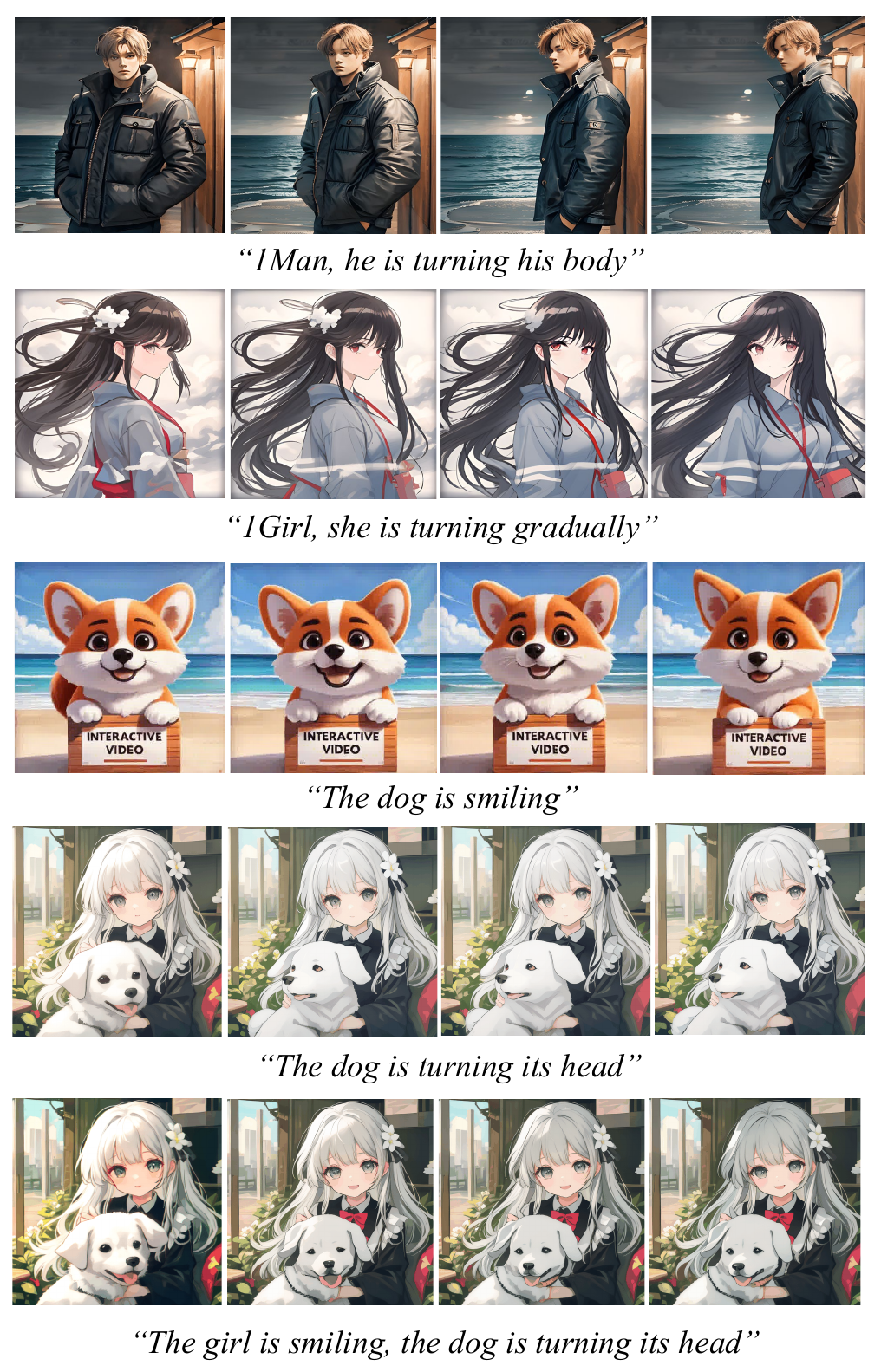}
    \caption{\textbf{Precise Motion Control} of \textit{InteractiveVideo}. Our framework shows strong controllability in large motion control, precise gesture control, and multi-object motion control.}
    \label{fig:5}
\end{figure}

\subsection{Precise Motion Control}~\label{sec:exp:4.3}

Motion control, particularly precise motion control, poses a significant challenge in the field of video generation due to the complexity of modeling spatial-temporal patterns. The primary difficulty lies in maintaining the temporal consistency of generated videos, especially when handling substantial motion. This issue mainly stems from the limited temporal receptive field of 1D temporal attention, which struggles to accommodate the full range of motion-related changes over time. As a result, ensuring smooth and consistent representation of motion in generated videos remains a considerable obstacle in this field. Differently, \textit{InteractiveVideo} excels in precise motion control, which we will discuss from three aspects as follows:

\textbf{1}) Large Motion. As shown in the first two rows of Figure~\ref{fig:5} first two rows, we present the large motion control by turning around characters in both realistic and cartoon styles. The details of turning around the female character are impressive, with the motion of her hair appearing highly realistic.

\textbf{2}) Precise Motion. As seen in the third row of Figure~\ref{fig:5}, the adorable corgi holding the ``INTERACTIVE VIDEO" brand displays several different charming gestures, including wagging its tail, smiling with an open mouth, turning its head, and shaking its ears.

\textbf{3}) Multi-Object Motion. The last two rows in Figure Figure~\ref{fig:5} showcase the ability of \textit{InteractiveVideo} to control multi-object motion. Our framework precisely controls the movements of both the cute girl and the lovable dog. When adjusting the dog's head, its tail also wags, and the girl naturally lowers her hand. While controlling these two objects, the girl smiles sweetly, and the dog turns its head to the other side.

\subsection{Quantitative Analysis}
\label{sec:exp:quan}

\paragraph{AnimateBench.} 
Since \textit{InteractiveVideo} is a general framework for open-domain video generation, we use \textbf{AnimateBench} for comparison. We assessed the text-based video generation capability using 105 unique cases with varying content, styles, and concepts. These cases were created using seven distinct text-to-image models, with five images per model for thorough comparison. Additionally, we crafted three motion-related prompts for each image to evaluate motion controllability across different methods, focusing on potential single-shot image motions.

\paragraph{Evaluation Metrics.} 
We evaluate generation quality by considering image and text alignment, using CLIP scores to measure cosine similarity between embeddings. Image alignment compares input images and video frame embeddings, while text alignment examines text and frame embedding similarities.

\begin{table}[ht]
\begin{center}
    \resizebox{0.93\linewidth}{!}{
    \begin{tabular}{lccccc}
        \toprule
        \multicolumn{1}{c}{\multirow{2}{*}{Methods}} & \multicolumn{2}{c}{CLIP Score} & \multicolumn{2}{c}{User Study} & Satisfaction Rate                                \\
        \multicolumn{1}{c}{}                         & Image                          & Text                           & Image          & Text      & (\%)     \\
        \hline
        VideoComposer\cite{wang2023videocomposer}    & 225.3                          & 62.85                          & 0.180          & 0.110  & 43.5        \\
        AnimateDiff\cite{guo2023animatediff}         & 218.0                          & 63.31                          & 0.295          & 0.220         & 51.6 \\
        PIA~\cite{zhang2023pia}  & {225.9}                 & {63.68}                 &{0.525} &{0.670}  & 52.5\\ \hline
        \textit{InteractiveVideo}~\pub{Ours} & \textbf{234.6} & \textbf{65.31} & \textbf{0.745} & \textbf{0.813} & \textbf{72.8}\\
        \bottomrule
    \end{tabular}
    }
\end{center}
\vspace{-5mm}
\caption{Quantitative comparison on AnimateBench.}
\label{tab:quant}
\end{table}

\paragraph{User Study.} To substantiate the enhancement of our method in terms of visual quality and user experience, we carried out a user study comparing our approach with other video models. This study utilized 40 prompts from AnimateBench, which feature a variety of scenes, styles, and objects. Compared to existing video generation methods, our \textit{InteractiveVideo} notably outperforms in terms of human preference scores and delivers state-of-the-art performance in user satisfaction rates. These quantitative results, coupled with the user study, effectively demonstrate the significance and superiority of the interactive generation paradigm and user-centric designs.

\subsection{Generation Efficiency}~\label{sec:exp:efficiency}

\textit{InteractiveVideo} takes only 16GB CUDA memory in the inference process, and it runs on a single RTX 4090. Besides, in Table~\ref{tab:effieciency}, we also report the latency of \textit{InteractiveVideo}. It is worth noting that \textit{InteractiveVideo} can generate a video within about 12 seconds though it requires two independent diffusion models for better controllability.

\begin{table}[ht]
    \centering
    \resizebox{0.97\linewidth}{!}{
    \begin{tabular}{lccccc}
        \toprule
         Process & Image Instruction & Content Instruction & Motion Instruction &Trajectory Instruction  \\ \hline
         Time & 19.34ms & 31.47ms &12.22s &77.35 ms \\
         \bottomrule
    \end{tabular}
    }
    \caption{Latency Analysis of \textit{InteractiveVideo}.}
    \label{tab:effieciency}
\end{table}

\section{Responsible AI and Ethic Claim}
In developing \textit{InteractiveVideo}, our research rigorously adheres to the principles of Responsible AI and ethical guidelines. This innovative framework for video generation is designed with a strong commitment to ethical AI practices, ensuring that user interactions with the system - through text, images, and direct manipulation - are processed with the utmost integrity and transparency. The implementation of our Synergistic Multimodal Instruction mechanism is a testament to our dedication to these principles. It not only facilitates a seamless integration of diverse user inputs but also ensures that the AI operates within ethical boundaries, avoiding biases and respecting user intent. By empowering users to interactively manipulate the video generation process, \textit{InteractiveVideo} promotes not just creativity but also responsibility in AI use. This approach aligns with our commitment to uphold ethical standards in AI, ensuring that \textit{InteractiveVideo} serves as a model for responsible innovation in the realm of AI-driven content creation

\section{Conclusion and Discussion}~\label{sec:conclusion}

In summation, we introduce \textit{InteractiveVideo}, a novel paradigm shift in the domain of video generation that champions a user-centric approach over the conventional methodologies reliant on pre-defined images or textual prompts. This framework is distinguished by its capacity to facilitate dynamic, real-time interactions between the user and the generative model, enabled by a suite of intuitive interfaces including, but not limited to, text and image prompts, manual painting, and drag-and-drop capabilities. Central to our framework is the innovative Synergistic Multimodal Instruction mechanism, a testament to our commitment to integrating multifaceted user interaction into the generative process cohesively and efficiently. This mechanism augments the interactive experience and significantly refines the granularity with which users can influence the generation outcomes. The resultant capability for users to meticulously customize key video elements to their precise preferences, coupled with the consequent elevation in the visual quality of the generated content, underscores the transformative potential of \textit{InteractiveVideo} in the landscape of video generation technologies.

\noindent\textbf{Discussion on the Computational Efficiency}. Notwithstanding the promising advancements heralded by \textit{InteractiveVideo}, the adoption of a user-centric generative approach is not devoid of challenges. Paramount among these is the imperative to ensure the framework's accessibility and intuitive usability across a broad spectrum of users, alongside maintaining the generative models' efficacy and computational efficiency amidst diverse and dynamic input scenarios. Future research endeavors might fruitfully focus on the refinement of these models to enhance scalability and the development of adaptive algorithms capable of more accurately interpreting and actualizing user intentions.

\noindent\textbf{Future Works}. We may delve into several promising directions. Enhancing the AI's understanding of complex user inputs, such as emotional intent or abstract concepts, could lead to more nuanced and contextually relevant video generation. Additionally, exploring the integration of real-time feedback loops where the model suggests creative options based on user input history could further personalize the user experience. Investigating the application of this framework in virtual and augmented reality environments opens up new dimensions for immersive content creation. Furthermore, extending the framework's capabilities to include collaborative generation where multiple users can interact and contribute to a single generative process may revolutionize co-creation in digital media.

\noindent\textbf{Further Applications}. The potential applications of \textit{InteractiveVideo} extend well into the realms of education, where bespoke video content could significantly enrich the pedagogical experience and entertainment, particularly in the creation of interactive narratives. As we continue to iterate upon and enhance this framework, the scope for its application appears limitless, heralding a future in which video generation transcends mere content creation to become a conduit for deep, interactive engagement between creators and their digital canvases.

\clearpage
\appendix
\noindent\textbf{\Large Appendix}

\begin{strip}\centering
	\vspace{-19mm}
	\includegraphics[width=0.80\textwidth]{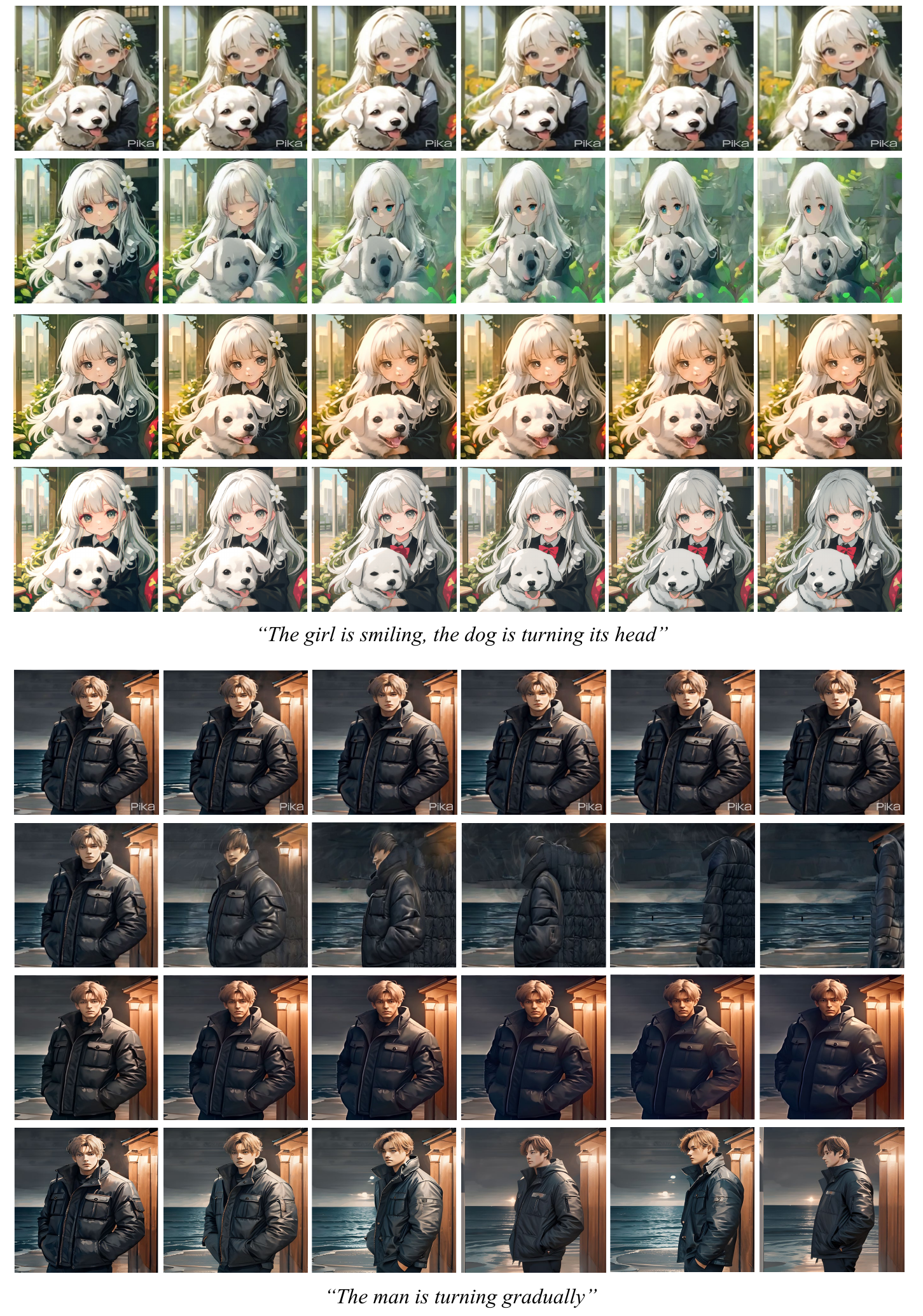}
 \vspace{17mm}
 \captionof{figure}{\textbf{Comparison with existing methods on motion control}. We compare \textit{InteractiveVideo} (4th row) with Pika Labs (1st row), I2VGen-XL (2nd row), and Gen-2 (3rd row).
 \label{fig:6}}
\end{strip}

\begin{strip}\centering
	\vspace{-19mm}
	\includegraphics[width=0.84\textwidth]{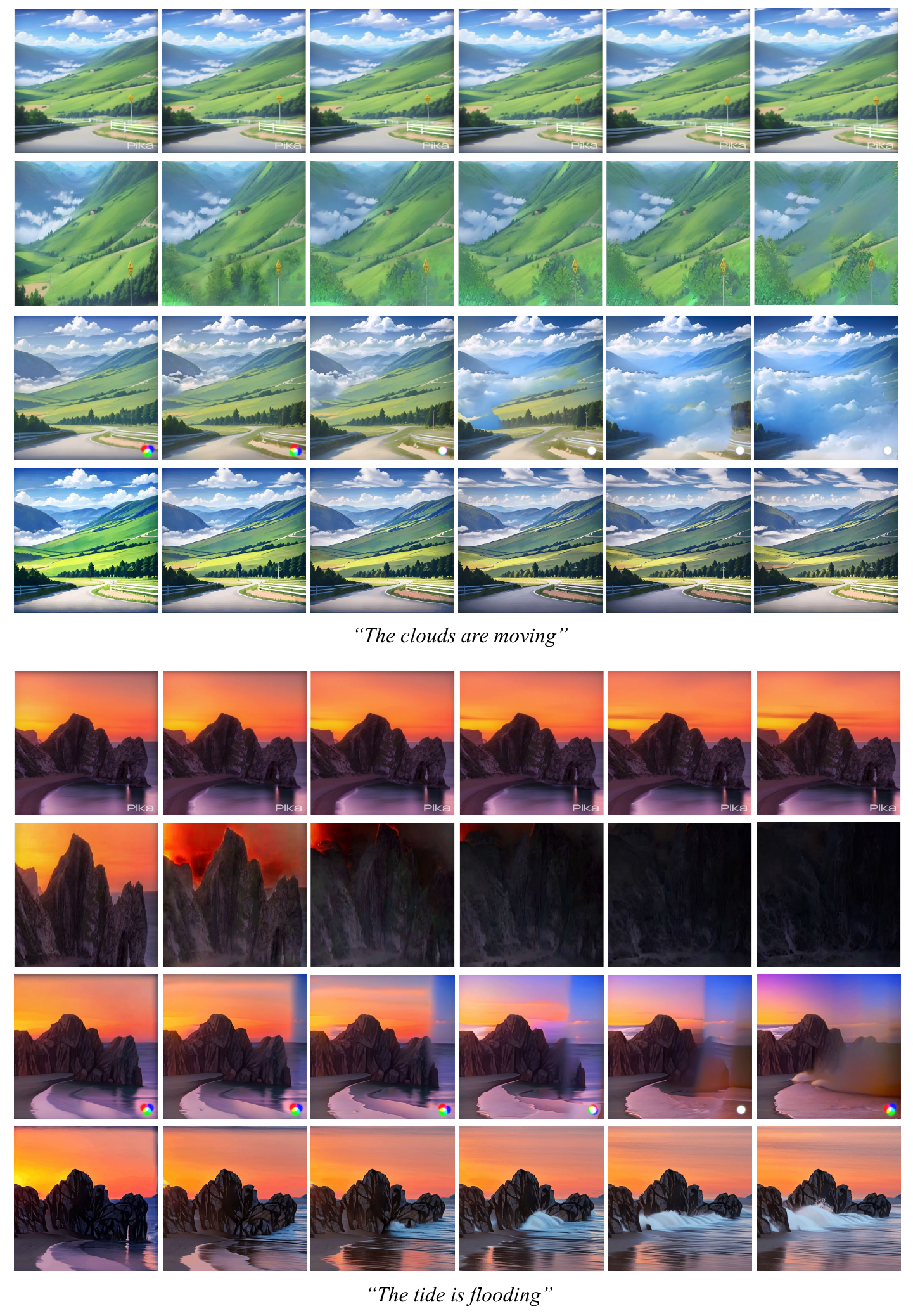}
 \vspace{17mm}
 \captionof{figure}{\textbf{Comparison with existing methods on landscapes}. We compare \textit{InteractiveVideo} (4th row) with Pika Labs (1st row), I2VGen-XL (2nd row), and Gen-2 (3rd row).
 \label{fig:7}}
\end{strip}

\begin{figure*}[ht]
\centering
\includegraphics[width=0.78\textwidth]{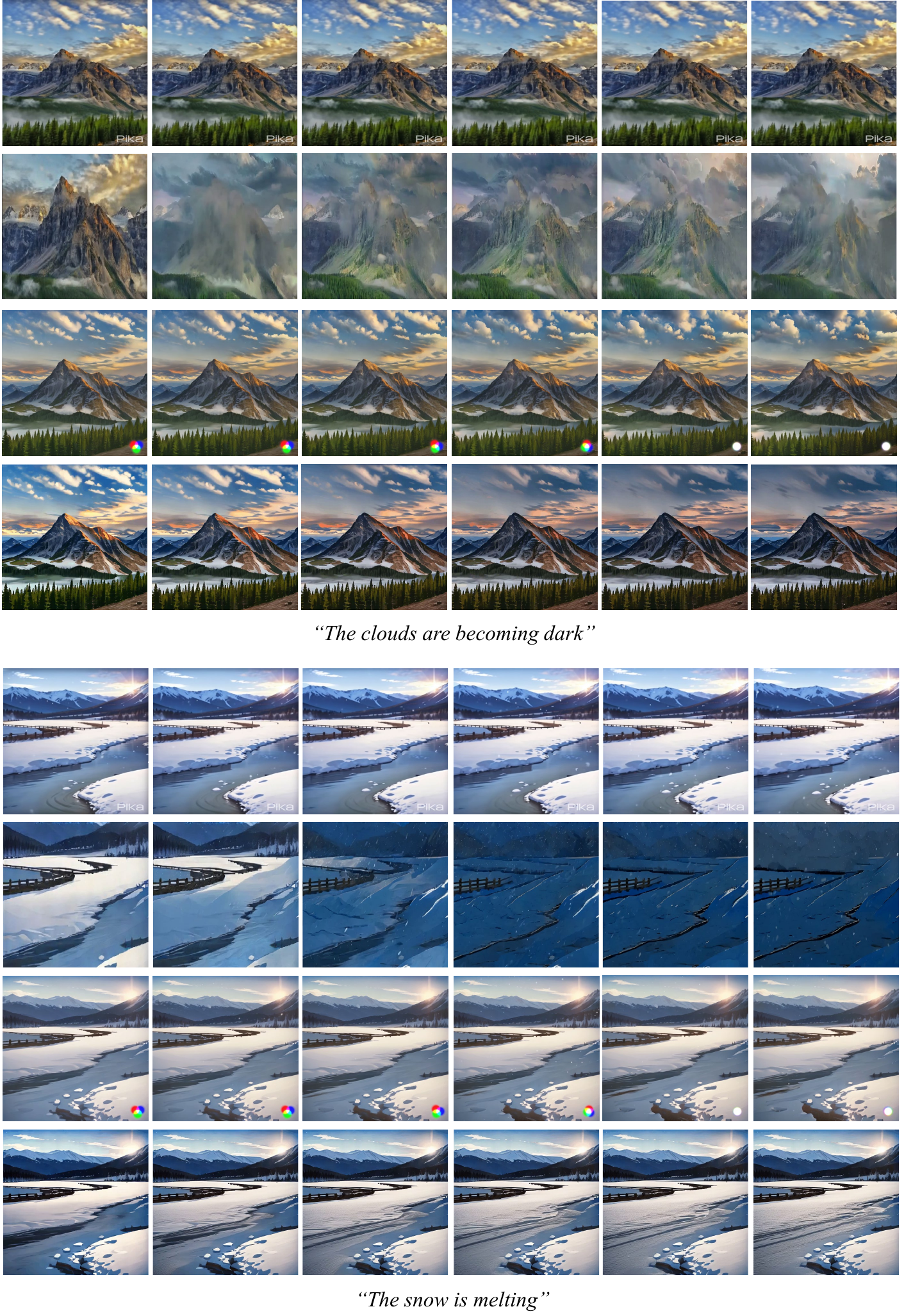}
 \captionof{figure}{\textbf{Comparison with existing methods on dynamic scenes}. We compare \textit{InteractiveVideo} (4th row) with Pika Labs (1st row), I2VGen-XL (2nd row), and Gen-2 (3rd row).
 \label{fig:8}}
\end{figure*}

\clearpage
{
	\small
	\bibliographystyle{ieeenat_fullname}
	\bibliography{main}
}



\end{document}